\documentclass[a4paper,conference]{IEEEtran}
\ifCLASSINFOpdf
\else
\fi
\usepackage{times}
\usepackage{epsfig}
\usepackage{graphicx}
\usepackage{amsmath}
\usepackage{amssymb}

\usepackage{graphicx}
\usepackage{epsfig}
\usepackage{epic,eepic}
\usepackage{times}
\usepackage{mathptmx}
\usepackage{amsfonts}
\usepackage{amsmath}
\usepackage{cite}
\usepackage{color}

\usepackage{times}

\definecolor{red}{rgb}{1,0,0}
\definecolor{green}{rgb}{0,1,0}
\definecolor{blue}{rgb}{0,0,1}
\definecolor{violet}{rgb}{1,0,1}
\definecolor{cyan}{cmyk}{1,0,0,0}
\definecolor{magenta}{cmyk}{0,1,0,0}
\definecolor{yellow}{cmyk}{0,0,1,0}

\definecolor{white}{rgb}{1,1,1}

\usepackage{jumoline}

\usepackage{arydshln}

\newcommand{\CO}[1]{}

\newcommand{\AC}{
\section*{Acknowledgment}

Our work has been supported in part by 
JSPS KAKENHI 
Grant-in-Aid 
for Scientific Research (C) 26330297,
and for Scientific Research (C) 17K00361.

}

\newcommand{\CommentOut}[1]{}

\newcommand{\noeditage}[1]{#1} \newcommand{\editage}[1]{}

\begin{document}

\newcommand{\FIG}[3]{
\begin{minipage}[b]{#1cm}
\begin{center}
\includegraphics[width=#1cm]{#2}\\
{\scriptsize #3}
\end{center}
\end{minipage}
}

\newcommand{\FIGU}[3]{
\begin{minipage}[b]{#1cm}
\begin{center}
\includegraphics[width=#1cm,angle=180]{#2}\\
{\scriptsize #3}
\end{center}
\end{minipage}
}

\newcommand{\FIGm}[3]{
\begin{minipage}[b]{#1cm}
\begin{center}
\includegraphics[width=#1cm]{#2}\\
{\scriptsize #3}
\end{center}
\end{minipage}
}

\newcommand{\FIGR}[3]{
\begin{minipage}[b]{#1cm}
\begin{center}
\includegraphics[angle=-90,clip,width=#1cm]{#2}
\\
{\scriptsize #3}
\vspace*{1mm}
\end{center}
\end{minipage}
}

\newcommand{\FIGRpng}[5]{
\begin{minipage}[b]{#1cm}
\begin{center}
\includegraphics[bb=0 0 #4 #5, angle=-90,clip,width=#1cm]{#2}\vspace*{1mm}
\\
{\scriptsize #3}
\vspace*{1mm}
\end{center}
\end{minipage}
}

\newcommand{\FIGpng}[5]{
\begin{minipage}[b]{#1cm}
\begin{center}
\includegraphics[bb=0 0 #4 #5, clip, width=#1cm]{#2}\vspace*{-1mm}\\
{\scriptsize #3}
\vspace*{1mm}
\end{center}
\end{minipage}
}

\newcommand{\FIGtpng}[5]{
\begin{minipage}[t]{#1cm}
\begin{center}
\includegraphics[bb=0 0 #4 #5, clip,width=#1cm]{#2}\vspace*{1mm}
\\
{\scriptsize #3}
\vspace*{1mm}
\end{center}
\end{minipage}
}

\newcommand{\FIGRt}[3]{
\begin{minipage}[t]{#1cm}
\begin{center}
\includegraphics[angle=-90,clip,width=#1cm]{#2}\vspace*{1mm}
\\
{\scriptsize #3}
\vspace*{1mm}
\end{center}
\end{minipage}
}

\newcommand{\FIGRm}[3]{
\begin{minipage}[b]{#1cm}
\begin{center}
\includegraphics[angle=-90,clip,width=#1cm]{#2}\vspace*{0mm}
\\
{\scriptsize #3}
\vspace*{1mm}
\end{center}
\end{minipage}
}

\newcommand{\FIGC}[5]{
\begin{minipage}[b]{#1cm}
\begin{center}
\includegraphics[width=#2cm,height=#3cm]{#4}~$\Longrightarrow$\vspace*{0mm}
\\
{\scriptsize #5}
\vspace*{8mm}
\end{center}
\end{minipage}
}

\newcommand{\FIGf}[3]{
\begin{minipage}[b]{#1cm}
\begin{center}
\fbox{\includegraphics[width=#1cm]{#2}}\vspace*{0.5mm}\\
{\scriptsize #3}
\end{center}
\end{minipage}
}

\onecolumn

\newcommand{\acprPaperID}{25}





\title{Use of Generative Adversarial Network for Cross-Domain Change Detection}

\author{\IEEEauthorblockN{Yamaguchi Kousuke ~~~~ Tanaka Kanji ~~~~ Sugimoto Takuma}
\IEEEauthorblockA{Graduate School of Engineering, University of Fukui\\
3-9-1, bunkyo, fukui, fukui\\
Email: tnkknj@u-fukui.ac.jp}
\thanks{Our work has been supported in part by 
JSPS KAKENHI 
Grant-in-Aid
for Scientific Research (C) 26330297,
and for Scientific Research (C) 17K00361.}
}

\newcommand{\figA}{
\begin{figure}[t]
  \begin{center}
\FIG{8}{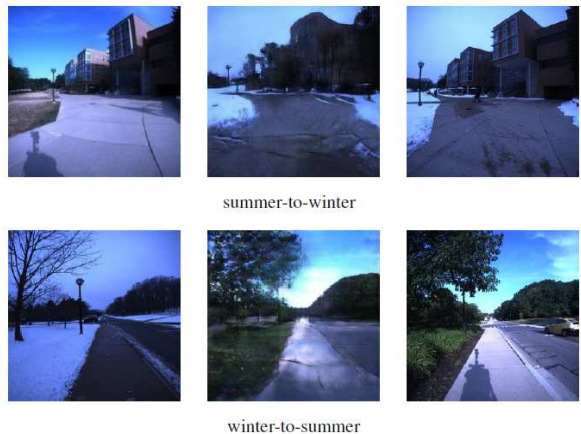}{}
\caption{
In our approach,
a generative adversarial network (GAN) -based
image translator
is trained and
used to map a reference image (left panel)
to a virtual one (middle panel) 
that cannot be discriminated from the query domain images.
This enables
us to treat the cross-domain change detection
as an in-domain image comparison
between
the query (right panel) and the virtual reference (middle panel) images.
}\label{fig:A}
\end{center}
\vspace*{-5mm}
\end{figure}
}

\newcommand{\figC}{
\begin{figure}[t]
  \begin{center}
\FIGR{8}{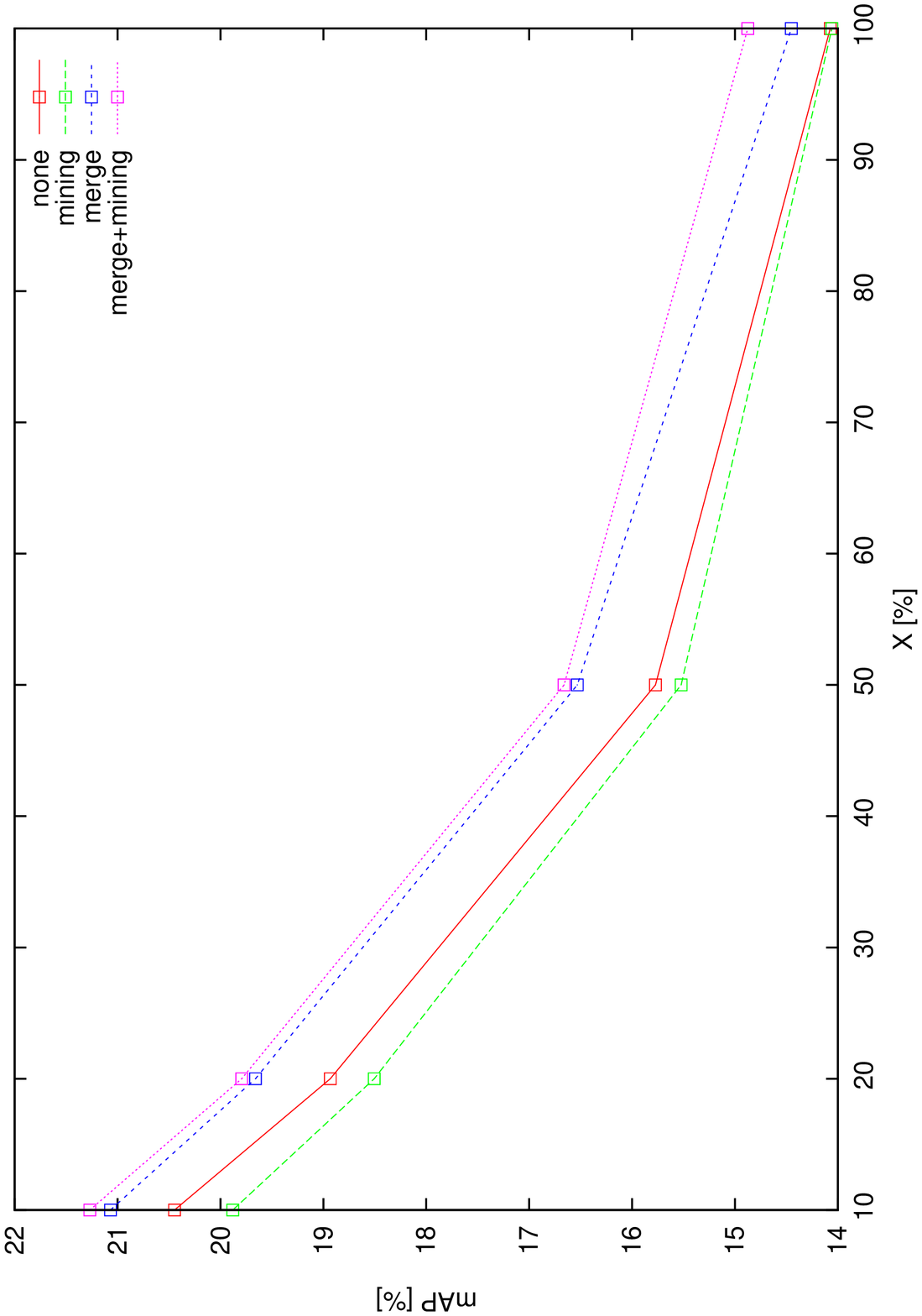}{b}
\caption{mAP@$X$[\%]recall performance
for different strategies.}\label{fig:C}
\end{center}
\vspace*{-5mm}
\end{figure}
}

\newcommand{\figD}{
\begin{figure*}[t]
  \begin{center}
\FIG{8}{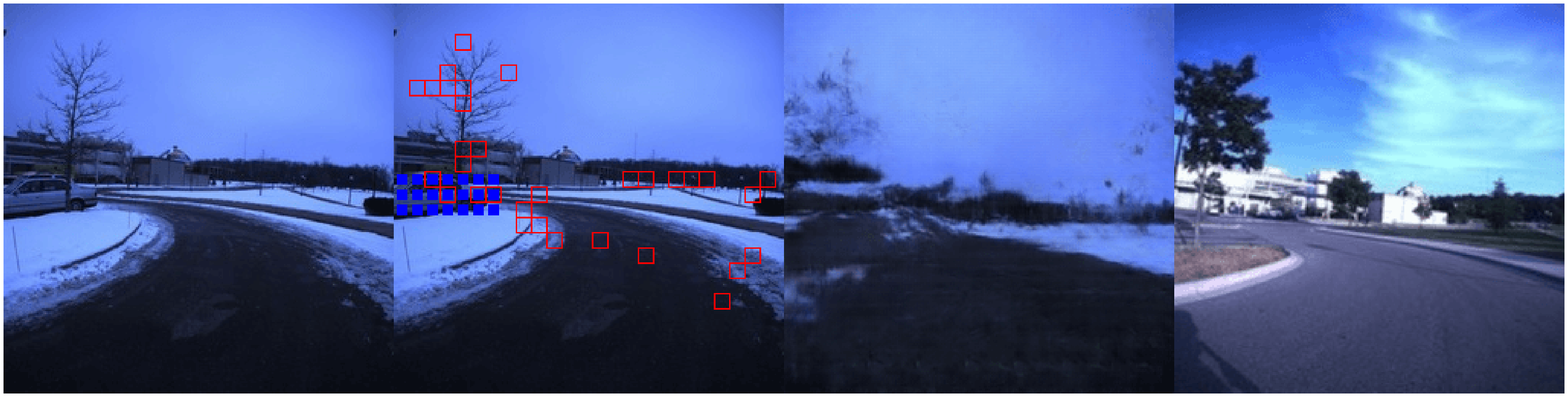}{a}
\FIG{8}{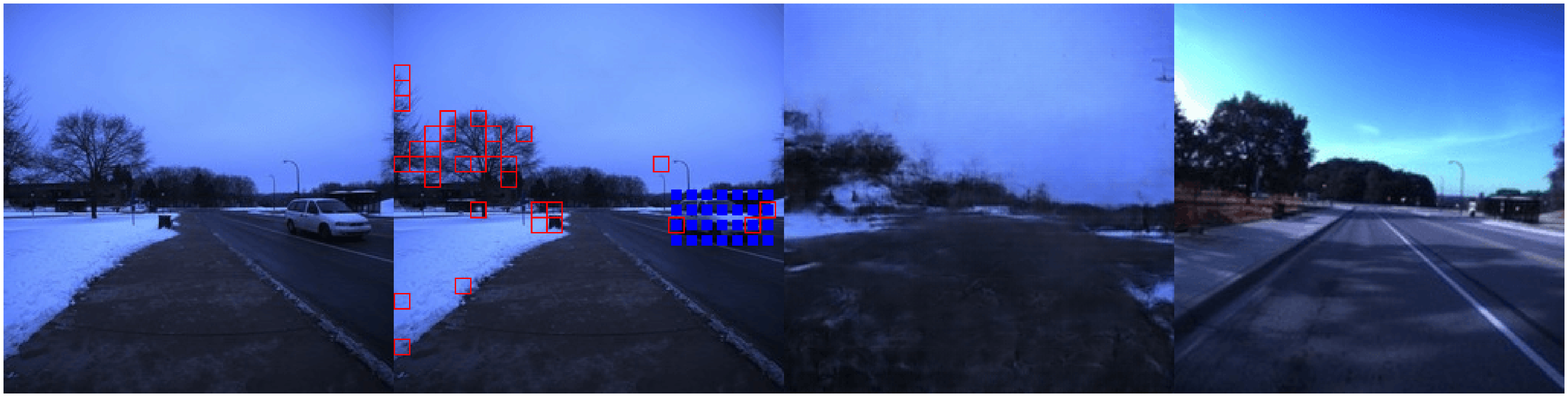}{b}\\
\FIG{8}{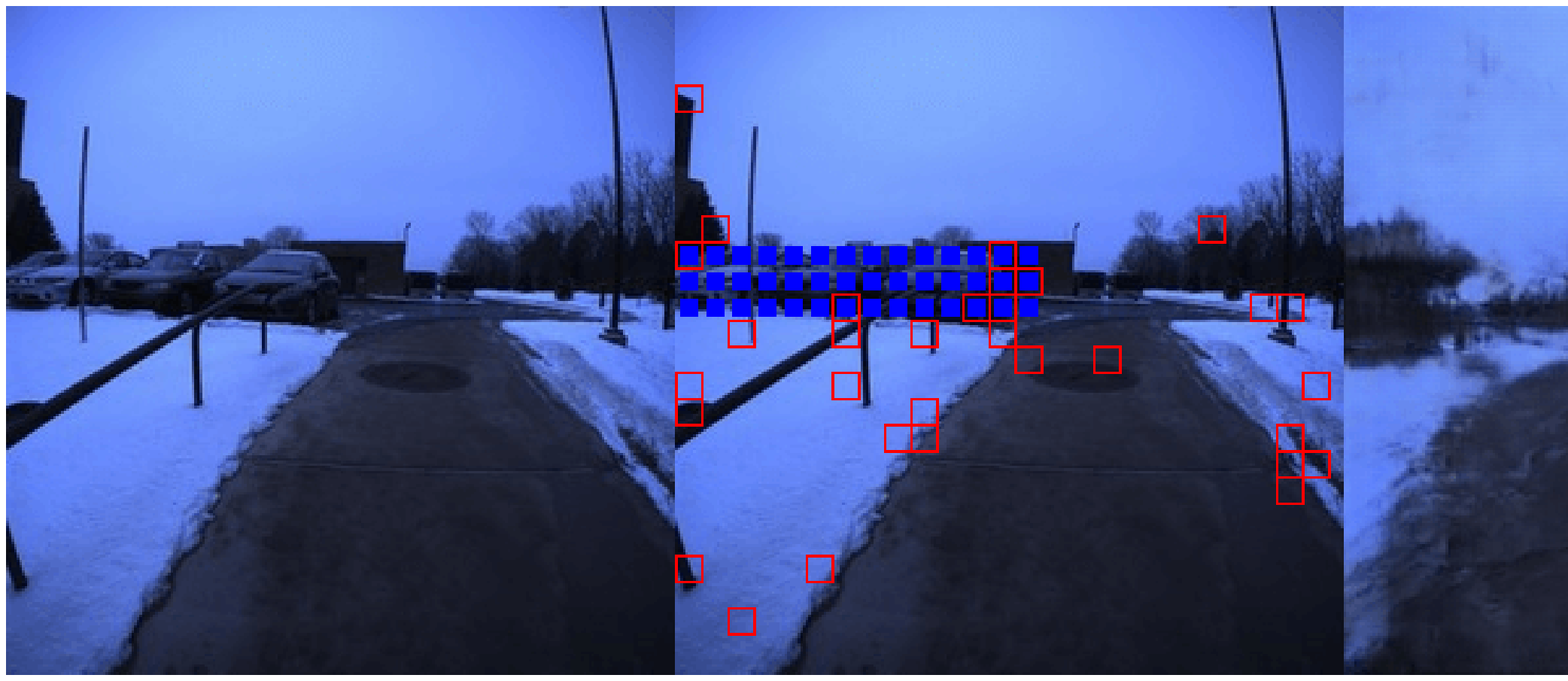}{c}
\FIG{8}{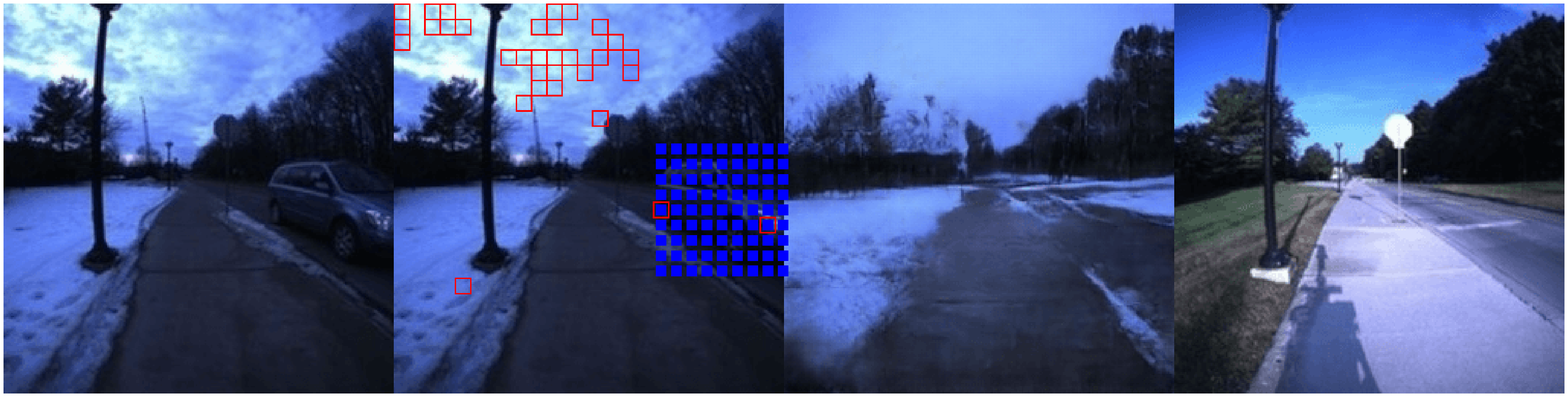}{d}\\
\FIG{8}{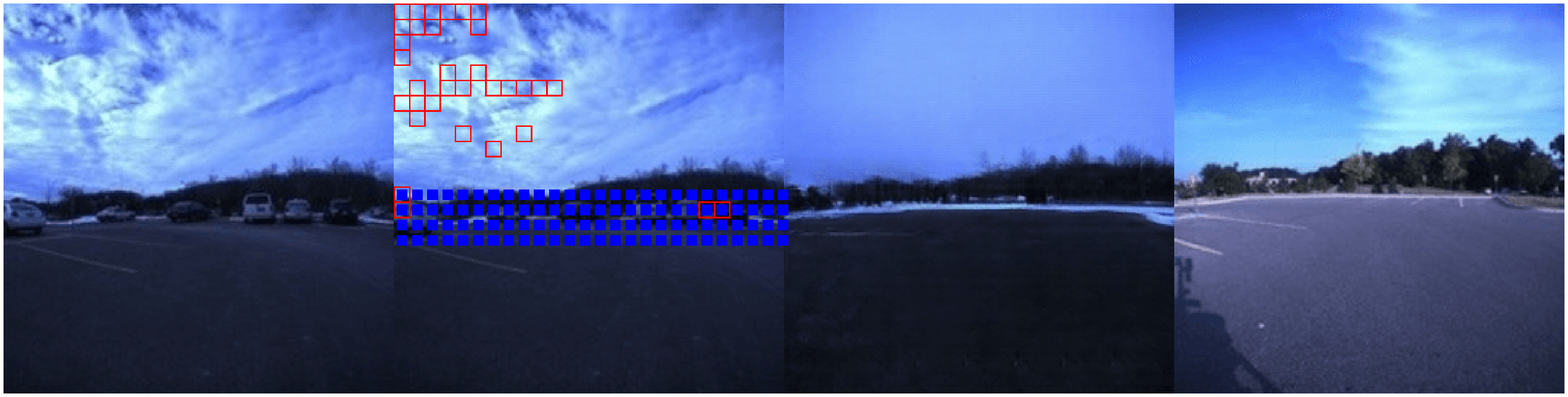}{e}
\FIG{8}{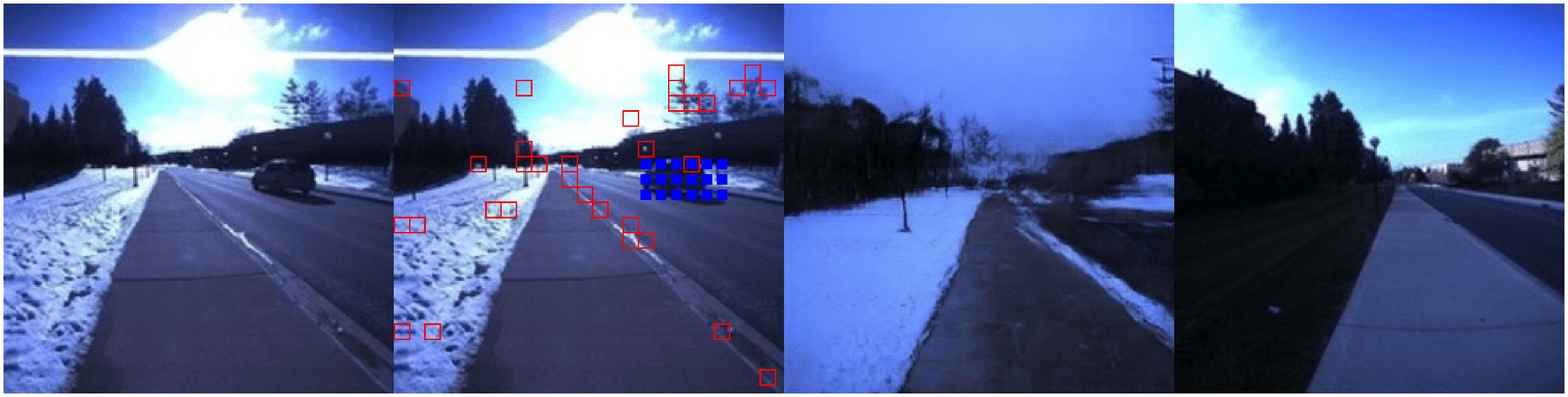}{f}\\
\FIG{8}{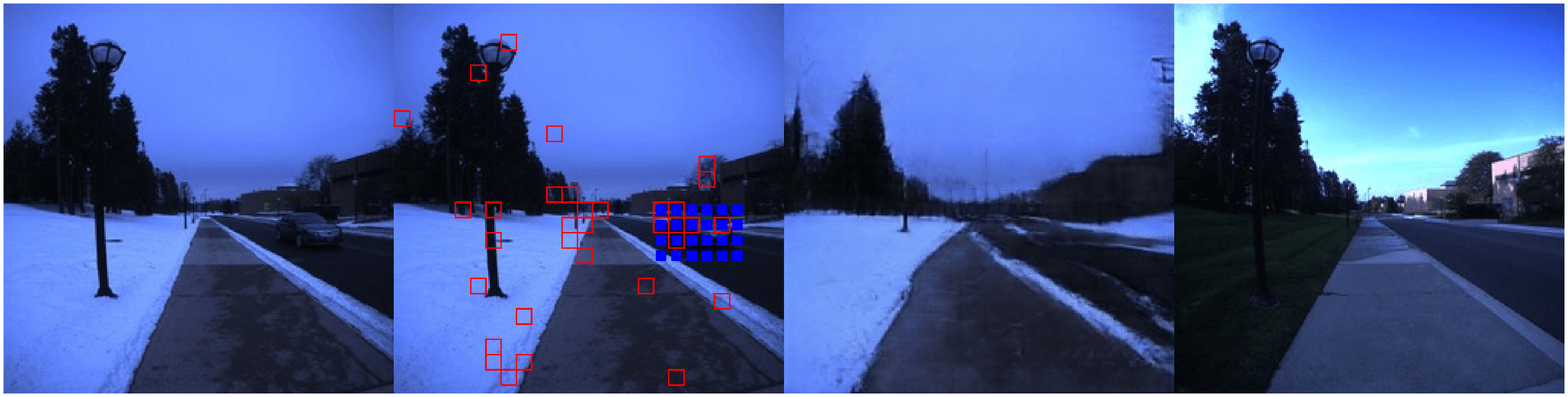}{g}
\FIG{8}{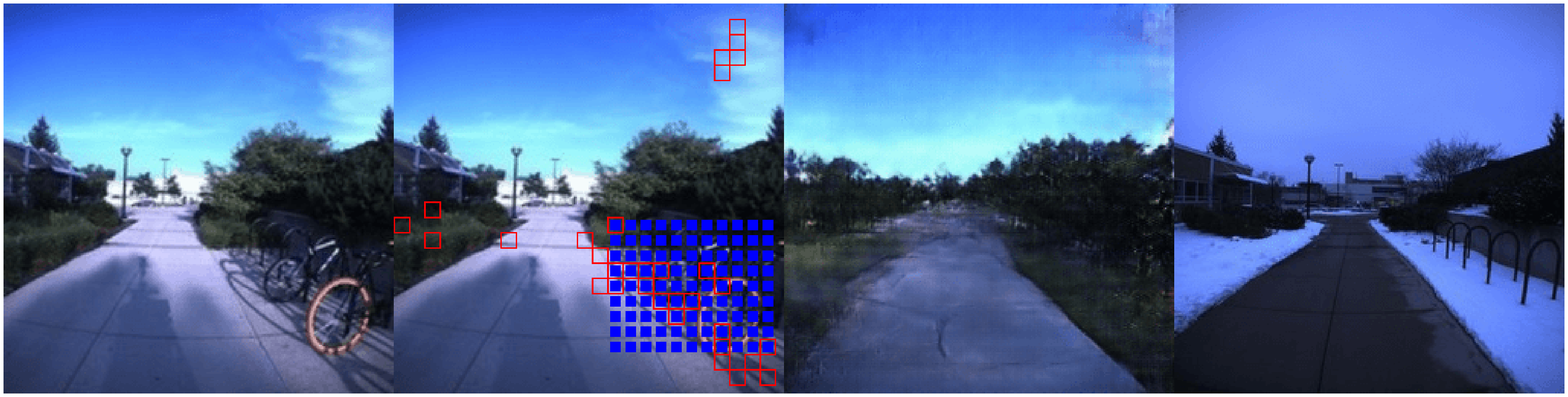}{h}\\
\FIG{8}{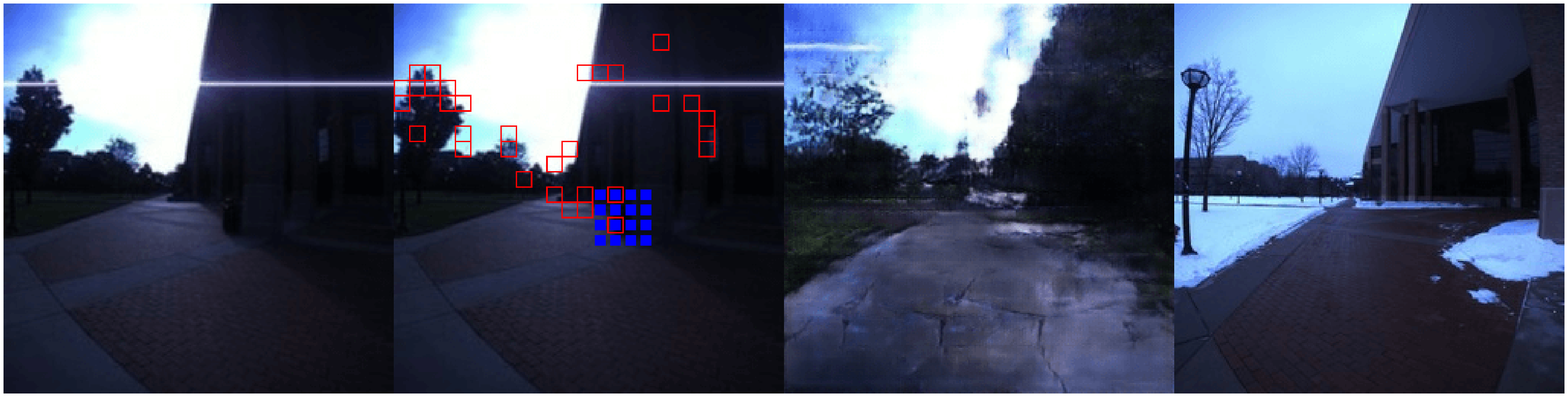}{i}
\FIG{8}{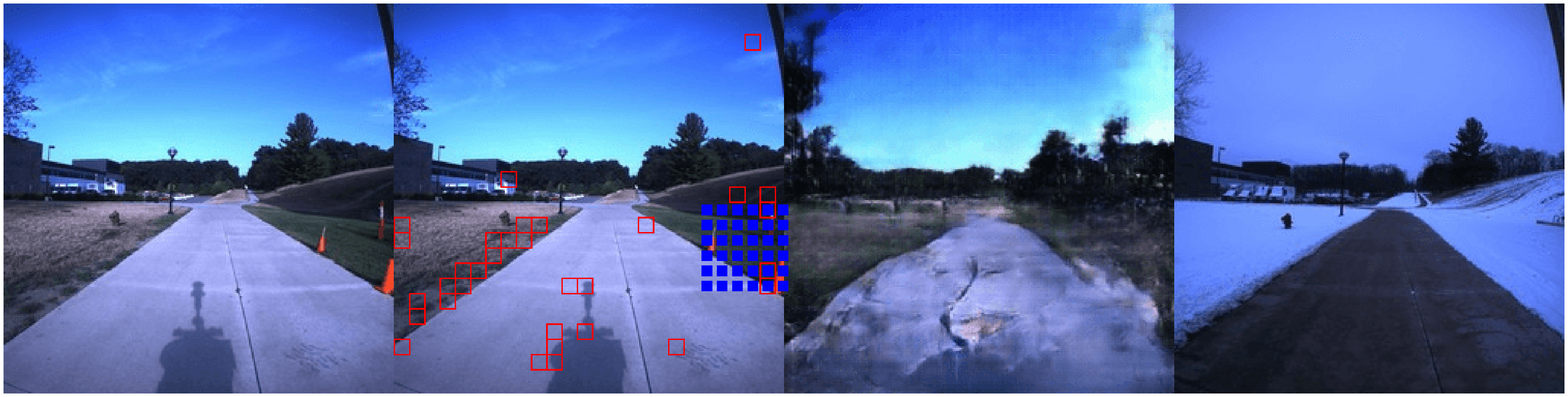}{j}\\
\caption{Examples of change detection.
From left to right,
each panel shows
the input query image,
prediction (red boxes) and ground-truth (blue points) change detection overlaid on the query image,
a virtual reference image,
and the input reference image.
The
prediction results (red boxes)
correspond to image grid cells with the top 5\% LOC values.
}\label{fig:D}
\end{center}
\vspace*{-7mm}
\end{figure*}
}

\newcommand{\figE}{
\begin{figure}[t]
  \begin{center}
\FIG{7}{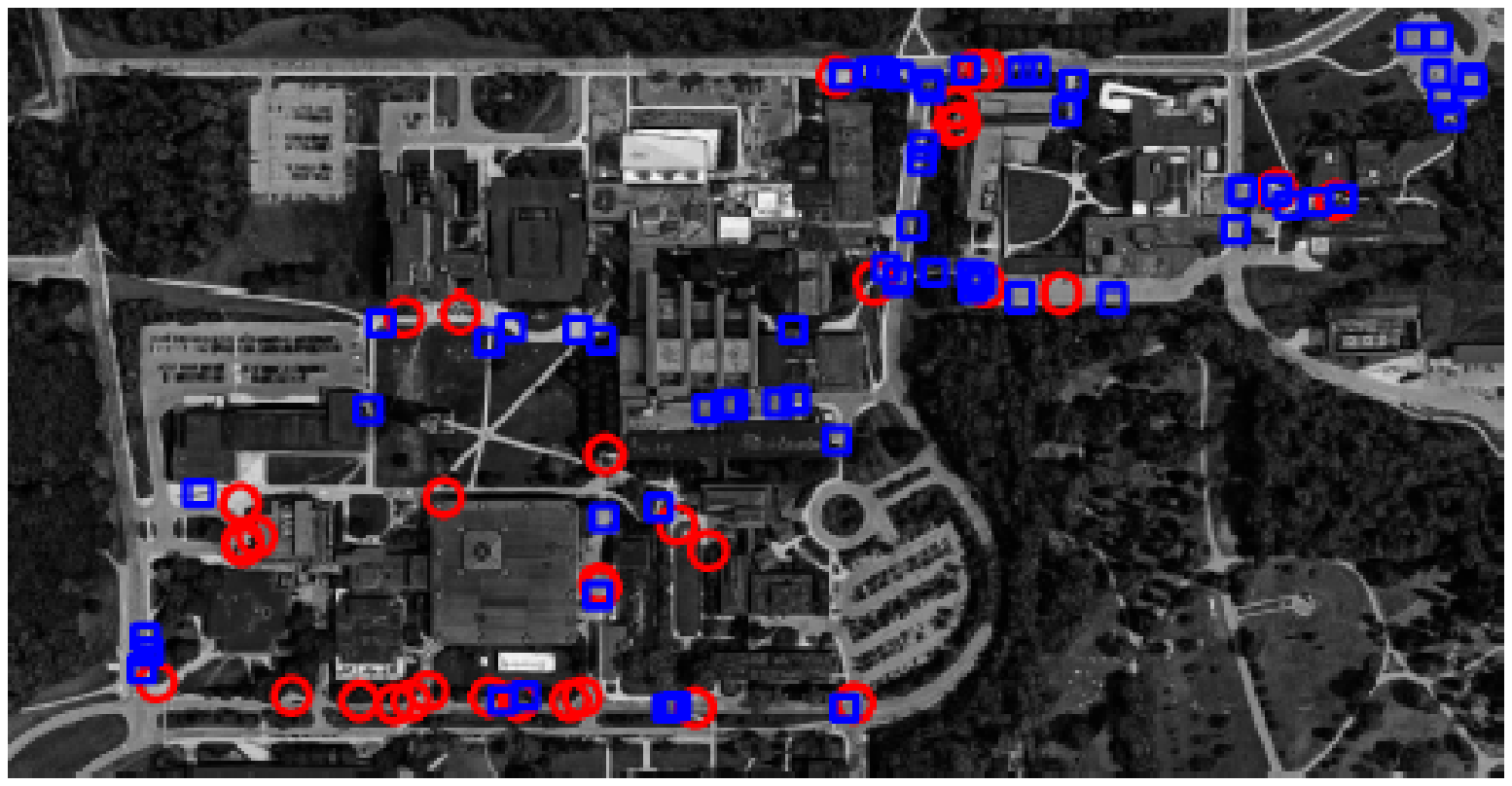}{}
\caption{Experimental environments
and viewpoints.
The red circles and blue boxes
indicate
the viewpoints of the 100 pairs of
summer and winter images
in the test set.
}\label{fig:E}
\end{center}
\vspace*{-3mm}
\end{figure}
}

\maketitle

\begin{abstract}
This paper addresses the problem of cross-domain change
detection from a novel perspective of image-to-image translation.
In general, change detection aims to identify interesting changes
between a given query image and a reference image of the same
scene taken at a different time. This problem becomes a challenging
one when query and reference images involve different domains
(e.g., time of the day, weather, and season) due to variations in object
appearance and a limited amount of training examples. In this study,
we address the above issue by leveraging a generative adversarial network
(GAN). Our key concept is to use a limited amount of training
data to train a GAN-based image translator that maps a reference
image to a virtual image that cannot be discriminated from query
domain images. This enables us to treat the cross-domain change
detection task as an in-domain image comparison. This allows us to
leverage the large body of literature on in-domain generic change
detectors. 
In addition, we also consider the use of visual place recognition 
as a method for mining more appropriate reference images over 
the space of virtual images. 
Experiments validate efficacy of the proposed approach.
\end{abstract}

\newcommand{\SW}[2]{#1}

\section{Introduction}

\SW{
This paper addresses the problem of change detection from a novel
perspective of domain adaptation. Change detection aims to identify
interesting changes between a given query image and a reference
image of the same scene taken at different times. This is a fundamental
problem in computer vision and robotics with many important applications
including visual navigation, novelty detection, surface inspection,
and city model maintenance. This problem becomes a challenging
one when query and reference images involve different domains (e.g.,
time of the day, weather, season). This is due to variations in object
appearance in different domains, which makes it harder to discriminate
changes of interest from nuisances. One of most basic schemes to
handle this difficulty is to train a domain-specific change predictor
for each possible domain pair. However, this requires a vehicle to
collect and learn from a number of training images from both domains,
which severely limits its applicability to new domains where large
amounts of annotated training data are typically unavailable.
}{
}

\noeditage{
\figA
}

\SW{
In this study, we address the above issue by leveraging a generative adversarial
network (GAN) \cite{A} (Fig. \ref{fig:A}). Our approach is motivated by the recent success
of the image-to-image translation technique in the field of computer
vision and graphics. The image-to-image translation technique aims to
learn the mapping between an input image in one domain to a virtual
image that cannot be discriminated from target domain images. The
key concept we employ is to use a limited amount of training data to train a GAN-based
image translator for reference-to-query domain translation. This
enables us to treat the cross-domain change detection task as an in-domain
image comparison and allows us to leverage the large body of
literature on in-domain generic change detectors. 
In addition, we also consider the use of visual place recognition 
as a method for mining more appropriate reference images over 
the space of virtual images. 
Experiments validate efficacy of the proposed approach.
}{
}

\section{Related Work}

\SW{
Change detection has been widely studied in various task scenarios.
These scenarios are categorized into 2D-to-2D matching \cite{B}, 
3D-to-3D matching \cite{C}, 
2D-to-3D matching \cite{D}. 
\cite{B} considers a 2D-to-2D matching on overhead imagery using
an image registration
and scale invariant feature transform (SIFT).
\cite{C} addressed a 3D-to-3D matching on point cloud data for novelty detection by patrol robots.
\cite{D} considered a 2D-to-3D matching between monocular image and cadastral 3D city models.
Our approach belongs to 2D-2D matching,
and our setting is
far more complicated than the case of overhead imagery \cite{B}
as our vehicular applications require the capability to deal with
more general 6 degrees-of-freedom (DoF)
freely-moving camera motion in a 3D space.

Up until now, the most fundamental scheme reported for addressing this challenge is to directly compare each view image against the corresponding reference image.
In \cite{F}, 
a scene alignment method for image
differencing is proposed based on ground surface reconstruction,
texture projection, image rendering, and registration
refinement. 
In \cite{G}, 
a deep deconvolutional network for pixelwise
change detection was trained and used for comparing
query and reference image patches.
However,
a major concern of these direct methods
is that they may fail to capture appearance variations
caused by a domain shift.
}{
}

\SW{
Our approach is inspired by the domain adaptation and transfer learning
approaches, ranging from parameter adaptation, feature transformation,
and metric learning, to deep learning techniques, which have
been applied to a wide variety of visual recognition tasks \cite{I}. 
In recent years, the application of GANs to domain
adaptation tasks has attracted much attention \cite{A}. In our application
domain, image-to-image translation has been one of mostly studied issues
over the last two years \cite{L}.  
Image-to-image translation aims to map
a raw image in a source domain to a virtual image in such a way that 
the virtual image cannot be discriminated from raw images in the
target domain. Typical training algorithms for the translation GAN take 
a set of paired source-target images 
as input. We address this basic
setting 
and use a typical architecture in 
\cite{L} for the current experimental system.
}{
}

\figE

\section{Problem: Cross-Domain Change Detection}\label{sec:prob}

\SW{
Although our approach is sufficiently general and applicable to various types of environments (indoor and outdoor for instance) and sensor modalities, in our experiments, we focus on the NCLT dataset \cite{P}. The NCLT dataset is a large scale, long-term autonomy dataset for robotics research collected at the University of Michigan's North Campus by a Segway vehicle robotic platform. The Segway was outfitted with: a Ladybug3 omnidirectional camera, a Velodyne HDL-32E 3D lidar, two Hokuyo planar lidars, an IMU, a single axis fiber optic gyro (FOG), a consumer grade global positioning system (GPS), and a real-time kinematic (RTK) GPS. The data we used in the research includes image and navigation data from the NCLT dataset. The image data is from the front facing camera (camera\#5) of the Ladybug3 omnidirectional camera. Fig. \ref{fig:E} shows a bird's eye view of the experimental environment and the vehicle's camera viewpoints.

During the vehicle's trips through the outdoor environments (Fig. \ref{fig:E}), it encounters various types of changes, which originate from the movement of individuals, parking of cars, building construction, and opening or closing of doors. There are also nuisances that originate from illumination alterations, viewpoint-dependent changes of objects' appearances and occlusions, weather variations, and falling leaves and snow. A critical and significant challenge in a substantial majority of change detection tasks is to discriminate changes of interest from nuisances. This renders our change detection task significantly more demanding.
}{
}

\SW{
Precision-recall is a standard performance index for image based change detection algorithms \cite{Q}.
As a performance index,
we use mean average precision (mAP),
derived from the field of visual object detection \cite{K}.
A change detection algorithm outputs
a prediction of
the likelihood of change (LOC)
for each cell on a
$w\times h$ image grid of LOC values,
as we will describe in \ref{sec:comparison}.
For the evaluation,
the LOC value of each cell in the image grid is thresholded into a state of change or no-change, and then the binary change mask is compared against a ground-truth change mask. Different threshold values provide different tradeoff points between precision and recall. The mAP can be viewed as a summary of these different tradeoff points and approximates the value of receiver operating characteristic (ROC). Following the literature \cite{K}, we approximate the average precision (AP) based on the average of ``interpolated" precision values 
for 11 different recall values,
0, 10, ..., 100 [\%]. The mAP is obtained as the mean of the AP values over all the query images. Considering the fact that achieving 100\% recall is not necessarily required
in typical change detection applications, we also evaluate mAP@$X${\%}recall for different percentage points $X=$10, 20, 50, and 100.
}{
}

\section{Approach: Domain-Adaptive Change Detection}

\SW{
Our domain-adaptive change detection (DA-CD) framework consists of three distinct steps: image adaptation, image description and image comparison. Image adaptation aims to map a given reference (or background) image to
a virtual image, using the GAN-based image-to-image translation, in such a way
that the virtual reference image cannot be discriminated from raw images in the target domain.
Image description aims to extract
a set of image feature descriptors from
either image for image comparison.
Image comparison aims to predict a change mask by comparing local features between the query and reference images.
The three steps above are detailed in the following subsections.
}{
}

\subsection{Image Adaptation}

\SW{
In order to train a GAN that translates an image from the source domain to the corresponding virtual images in the target domain, we require a small number of paired source-target images as training data.
We pair images based on their viewpoint locations.
The task of pairing requires global viewpoint information in the source and target domains.
Fortunately, in vehicular applications, it is often straightforward to
obtain the small set of training images 
and the aforementioned global viewpoint information.
First, a small number of training sets
can be acquired by the vehicle's short range navigation in the target domain.
Second, 
if the GPS measurement of the viewpoints are available,
a target image with nearest neighbor viewpoint is selected as the corresponding image.
Even in a GPS-denied environment, pseudo GPS information can be obtained with sufficiently high quality by using 
a modern visual place recognition (VPR) technique. 
For example, in \cite{H}, we also have developed a robust approach to
cross-season VPR.
}{
}

\SW{
The GAN-based translation problem is often formulated as a per-pixel regression problem. However, this formulation treats the output space as ``unstructured" in the sense that each output pixel is considered conditionally independent from all others given the input image. This issue is addressed by conditional GAN with a structured loss function, which penalizes a joint configuration of the output. In particular, the approach proposed in \cite{L} 
aims for a general-purpose solution to image-to-image translation problems.
The proposed approach consists of
a generator based on U-Net 
which allows low-level information 
to take a shortcut across the network,
and a discriminator
based on PatchGAN 
that combines a typical 
L1 loss (that enforces correctness at low-frequencies) 
with an additional patch level loss (that models high-frequencies).
We also refer to the results in 
\cite{L}, 
where the approach was found to be effective for  a domain adaptation task of ``day-to-night" image translation,
which is partly 
similar to the application of a cross-season image translation task that we present.
In both the training and translation tasks,
we resize each input image to 256$\times$256.
}{
}

\subsection{Image Description}\label{sec:description}

\SW{
We follow the literature (e.g., \cite{R}) and use the standard local image feature descriptor, SIFT \cite{M}, as the basis for image comparison and change detection. 
In this study,
we are particulary interested in how effective intensity-non-invariant feature descriptors like SIFT descriptors are when combined with our strategy of image adaptation. Intuitively, when the image adaptation is successful, 
the image comparison between the query image and the adapted background image can be viewed as an in-domain image comparison task rather than a cross-domain image comparison. Therefore, intensity-non-invariant feature descriptors are expected to be effective for such a pseudo in-domain image comparison. 
Each SIFT feature descriptor is L1 normalized.
}{
}

\SW{
We basically use a Harris-Laplace detector as a feature keypoint
detector. This detector typically provides a set of well-localized
keypoints. However, in low texture image regions, it fails to detect
localized keypoints and the spatial density of keypoints tend to be very
low. To compensate, we also employ an alternative dense sampling 
-based detector. The dense sampling -based detector outputs a large
number of spatially uniform keypoints even in a low texture region.
However, as a downside, such keypoints from dense sampling tend to
be not well-localized and have a low invariance. This potentially yields
a bias due to an increase in dissimilarity between local features at these dense
keypoints, which then increases the false positive change detection. To deal
with this issue, we decided to penalize the LOC values of these features
by multiplying a pre-defined coefficient $\exp^{-d_a^2}$ ($d_a\gg 1$). 
}{
}

By default, the features from the virtual image are merged with those
from the raw reference image, and then the merged set is used for the
image comparison. In the experiments, we also consider an alternative
setting where the virtual image set alone is used (without the merge)
for image comparison. The above two strategies are respectively
termed ``w/ merge" and ``w/o merge".

\subsection{Image Comparison}\label{sec:comparison}

\SW{
The image comparison step adopts the nearest neighbor (NN) search
of local features. In offline mode, every feature in the reference image is
indexed by an NN data structure. Once online, the NN search is performed
for each query feature over the reference (raw/virtual) features. The
search result provides a dissimilarity score of a query feature's NN
reference features with respect to the query feature, which can be
interpreted to the LOC score of the query feature. We measure the
dissimilarity by calculating the Euclidean distance $|f_i-f_g|$ between query feature 
$f_i$ and its NN feature $f_g$. The search region on the image plane is
defined as a radius 10-pixel circular region centered at the query
keypoint of interest, 
for 256$\times$256 query and reference images.
}{
}

\SW{
We also adopt a feature matching -based 
feature penalization.
This strategy 
has been found to be effective in our previous research in \cite{N}. 
The basic function of the feature matching approach is to conduct a similarity search for a given query feature over a collection of spatially relevant reference features. If a strong match is found,
we can expect the actual possibility of the feature
being changed to be very low. Therefore, we  penalize the LOC values of these features by multiplying by  pre-defined coefficient $\exp^{-d_b^2}$ ($d_a\gg d_b\gg 1$). 
To detect strong matches,
we use the SIFT ratio test \cite{O} 
with a threshold value of 0.4. 
In preliminary experiments,
we also tested different threshold values ranging from 0.4 to 0.8 and found 
that the difference 
among these threshold values yield 
a low significance in change detection performance.
The feature matching approach is expected
to effectively suppress false positive change detection even
when ambiguous anomaly features exist.
}{
}

\SW{
In typical change detection applications, we are interested in pixel-level LOC, rather than feature-level LOC \cite{Q}.
To obtain such pixel-level LOC values, we introduce a 
coarse $w\times h$ image grid of LOC values,
where the width $w$ and height $h$ of the grid is set
to $\lfloor W/10 \rfloor$ and $\lfloor H/10 \rfloor$ for a $W\times H$ image.
The LOC value of each grid cell is computed from the LOC values of query features that belong to that grid cell by using max pooling.
}{
}

\SW{
In addition,
we also explored the use of the VPR technique as a method for selecting appropriate background images, which we call a ``background mining" strategy.
In general,
a virtual image generated by a
GAN-based image-to-image translation
is often inconsistent with the corresponding query image.
This is natural
because 
the query image is unseen and not available at the GAN training stage.
To address this issue,
here we use the VPR technique as a method for searching for a more relevant background image to a given query image,
over the space of all the virtual images of the target domain. In this paper, the VPR algorithm adopts a local feature NN search with a Naive Bayes nearest neighbor (NBNN) distance metric, which was found to be effective in a cross-domain VPR task in our previous research \cite{H}.
In experiments, we consider and compare two distinct cases: one for the normal background image output by the GAN-based image-to-image translator, and one for the virtual background image selected
by background mining.
}{
}

\figC

\figD

\section{Experiments}

\SW{
We evaluated the change detection strategies using an NCLT dataset \cite{P}. 
As described in \ref{sec:prob},
the NCLT dataset is a large scale, long-term autonomy dataset for robotics research collected at the University of Michigan's North Campus by a Segway vehicle robotic platform.
During the vehicle's trip through outdoor environments,
it encounters various types of interesting changes and nuisances.
A critical and significant challenge in a change detection task
is to discriminate changes of interest from nuisances.
We consider two different scenarios of domain-shift: winter-to-summer and summer-to-winter.
Our winter image set is generated from the datasets ``2012/1/22", ``2012/1/15", and ``2013/1/10" in NCLT,
while our summer set is generated from ``2012/8/4",
by the following procedure.
First, images from different seasons are paired using the GPS information available in the NCLT dataset. Then, a random subset of the paired images is used for training the GAN-based image-to-image translator. Two translators corresponding to winter-to-summer and summer-to-winter translation tasks are trained using 100 and 100 paired training images, which are independent from the test sets and randomly sampled from the both domains. Then, every paired image is manually examined and annotated with a bounding box of a change object. We obtain 100 annotated image pairs as the test set, which consist of 64 summer-to-winter pairs and 36 winter-to-summer pairs.
}{
}

\SW{
We categorized small changes (e.g., 10$\times$10 [pixels]) that typically
originate from distant objects, into no-change because it is challenging to detect such small changes via a visual change detection algorithm. As a result, such small objects are likely to cause pseudo false-positive detection by change detection algorithms.

The average time-overhead for image translation was
around 1.0 sec and the time-overhead for change detection was 1.0 sec on a single PC (Intel Core i7-7700 3.2 GHz, GeForce GTX 1080). 
}{
}

\SW{
Fig. \ref{fig:C}
shows the mAP@$X${\%}recall performance for different change detection strategies.
As mentioned in \ref{sec:description},
we consider two settings,
whether or not
features from the virtual reference images
are merged with
features from the raw reference image,
which
are respectively called ``w/ merge" and ``w/o merge".
In addition,
we consider alternative settings
and whether or not we use the background mining described in \ref{sec:comparison},
which is termed ``w/ mining" and ``w/o mining".
Figure \ref{fig:C}
shows the
mAP performance for 
different combinations of
these three strategies.
It can be observed
that
the combination ``merge+mining"
yields the highest mAP@100{\%}recall performance.
}{
}

Fig. \ref{fig:D}
displays examples of change detection results.
It can be seen that
in these examples,
the virtual reference image preserves
the color and structure information
and successfully generates a realistic virtual image,
while the edge information is often blurred or lost.
It is noteworthy that
even with such erroneous edge information,
the change detection was successful as reported in 
Fig. \ref{fig:C}.

\section{Conclusions}

\SW{
In this paper, we discussed the use of
a generative adversarial network (GAN)
in cross-domain change detection tasks.
We proposed a novel framework of
domain adaptive change detection (DA-CD)
to 
capture the domain shift in the images,
and combined the
local feature search with
the GAN-based image translation
to capture spatial inconsistencies
and appearance inconsistencies between different domains.
Improved detection performance was achieved
on the challenging cross-season NCLT dataset.
}{
}

\AC{}

\bibliographystyle{IEEEtran}
\bibliography{gan}

\end{document}